# Evaluation of GPT-3 for Anti-Cancer Drug Sensitivity Prediction

Shaika Chowdhury, PhD[1], Sivaraman Rajaganapathy, PhD[1], Lichao Sun, PhD[1,2],
James Cerhan, MD, PhD[1], Nansu Zong, PhD[1*]
[1]Mayo Clinic, Rochester, MN, USA; [2]Lehigh University, Bethlehem, PA, USA

**Introduction**

Cancer is a complex genetic disease that originates from the accumulation of gene mutations within a cell and is ranked as the second leading cause of death in the United States according to the American Cancer Society[1]. Given the tumor heterogeneity arising from the genetic variations among patients even with the same cancer type, substantial differences in the anti-cancer drug response can be expected, thereby highlighting the urgent need for targeted therapies. Owing to the high cost and time associated with developing and validating anti-cancer drugs in clinical trials which is further exacerbated by the 96% failure rate, the development of preclinical computational models that can accurately predict whether a cell line is sensitive or resistant to a particular drug is imperative. The availability of large-scale pharmacogenomics datasets collected via high-throughput screening technologies offers feasible resources to develop robust drug response models and identify the important biomarkers predictive of drug sensitivity.

Large language models (LLM), such as the Generative Pre-trained Transformer (GPT-3) from OpenAI, are "task-agnostic models" pre-trained on large textual corpora crawled from the Web that have exhibited unprecedented capabilities on a broad array of NLP tasks. Recent studies have noted the potential of GPT-3 in the biomedical domain[2,3]; however, these studies focus on processing NLP datasets that include unstructured text and their applicability to biomedical tasks with structured data (e.g., pharmacogenomics data) remains unexplored. To this end, this work aims to investigate GPT-3's potential for anti-cancer drug sensitivity prediction on the Genomics of Drug Sensitivity in Cancer (GDSC)[4] database containing tabular pharmacogenomic information. The main contributions of this work include: (1) task-specific prompt engineering of the structured data, (2) evaluating and comparing the performance of GPT-3 for drug sensitivity prediction in the zero-shot and fine-tuning settings, (3) analyzing the effect of simplified molecular input line entry specification (SMILES) sequences of drugs and genomic mutation features of cancer cell lines on the model's generalization and (4) we release a web app for using the GPT-3 variant fine-tuned on the GDSC data for drug sensitivity classification at https://huggingface.co/spaces/ShaikaChy/SensitiveCancerGPT.

**Methods**

*Dataset:* We utilized the drug-cancer cell lines pairs and their corresponding drug response data (i.e., the half maximal inhibitory concentration (IC50)) from the new version of GDSC database (GDSC2) across 5 tissue types - Lung adenocarcinoma (LUAD), Breast invasive carcinoma (BRCA), Colon and rectum adenocarcinoma (COREAD), Thyroid carcinoma (THCA) and Brain Lower Grade Glioma (LGG) – which in total cover 288 unique drugs and 183 unique cell lines. We created dataset per tissue type which resulted in 16378, 20372, 35138, 1132 and 2516 drug-cell line pairs for the LUAD, BRCA, COREAD, THCA and LGG cohorts, respectively; each cohort was trained and evaluated using GPT-3 separately with 80%-20% stratified split for the training and test sets. In addition, to inspect the effect of integrating additional context with the input in the form of drug's chemical structure (SMILE) and gene mutation information on the model's performance, we create ablated datasets wrt different input combinations for LUAD as the illustrative tissue type. The following are the ablated input combinations and their data sizes: drug + cell line + smile (12003), drug + cell line + mutation (4469), drug + cell line + smile + mutation (3500).

*Task Overview:* We formulate drug sensitivity prediction as a binary classification problem to predict if a drug-cell line pair is sensitive or resistant. To convert the IC50 drug response values to binary labels, we check if the 'feature delta mean IC50' value is negative (sensitive) or positive (resistant).

*Zero-Shot Prompting:* Considering the test set as a M x N table of M drug-cell line pairs as the rows and N feature columns (i.e., "drug name", "drug target", "drug smile", "gene mutation", "drug response"), we first convert the structured cell values in each row to a natural language text $T$ using the corresponding column names (e.g., "The drug name is pci-34051. The drug target is hdac1. The drug smile is COC1=CC=C(C=C1)CN2C=CC3=C2C=C(C=C3)C(=O)NO. The gene mutation is crebbp. Drug response:"). Note that the last column is left blank for the model to predict. We then prepare a task-specific instruction $I$ "Decide in a single word if the drug's response to the target is sensitive or resistant." and concatenate it with $T$ to get the final prompt *P*. The prompt *P* is directly inputted into the GPT-3 Ada model via OpenAI Completions API to generate the response *R* corresponding to the model's drug response prediction.

*Fine-Tuning:* We prepare the training and test data in the form of prompt-completion pairs. The prompt is similar to that used in zero-shot setting but more concise as we prepend the column names (e.g., 'drug name:', 'drug target:', 'gene mutation:') before the respective cell values and concatenate them together using new line as the delimiter (e.g., drug: pci-34051\ndrug target: hdac1\ngene mutation: crebbp). The completion is the ground truth sensitive/resistant label. We fine-tune GPT-3 Ada model on the training set for 4 epochs and evaluate performance on the test set.

**Results**

We first compare the performance of GPT-3 in the zero-shot (Figure 1 subplot (i)) and fine-tuning (Figure 1 subplot (ii)) settings evaluated on the five tissue type datasets. We observe that although zero-shot prompting outperforms the fine-tuning counterparts across all the tissue types in F1 generally, the per class performance for zero-shot is heavily skewed toward the 'sensitive' class as the F1 scores for the 'resistant' class (F1-Resistant) are comparatively very low. We then analyze the results for fine-tuning with different feature combinations from the LUAD dataset, as reported in Figure 1 subplot (iii). The results reveal that gene mutation features alone and also in combination with the drug's smile representations are more informative of drug response, with 24% and 29% performance gains in F1 respectively. Subsequently, the fine-tuning performance on the other tissue cohorts with the best performing feature combination (i.e., Drug + Cell line + Smile + Mutation) is summarized in Figure 1 subplot (iv).

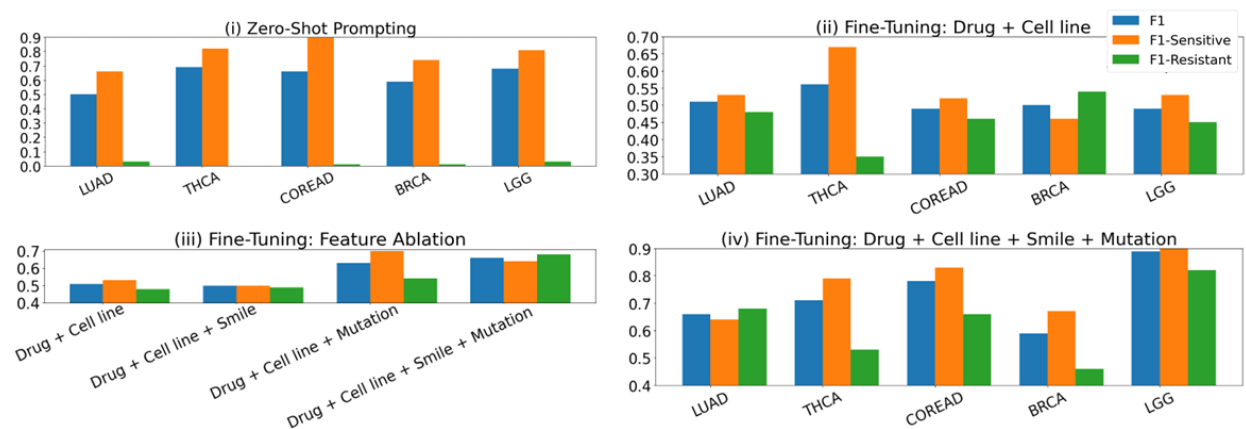

**Figure 1.** Performance evaluation of GPT-3 on different tissue types in (i) zero-shot setting, (ii) fine-tuning using Drug + Cell line, (iv) Drug + Cell line + Smile + Mutation as the input (iii) fine-tuning on LUAD with feature ablation.

**Discussion**

This study yields encouraging results for anti-cancer drug sensitivity prediction by employing generative large language model on structured pharmacogenomics data and offers a new direction in AI-based precision oncology. Comparative analysis was performed to demonstrate GPT-3's drug response generalizability in the zero-shot vs fine-tuning settings, where the fine-tuning performance was further enhanced with the use of gene mutation and drug's smile features. We believe the per-class performance with zero-shot prompting could potentially be improved through the pre-training of GPT-3 on biomedical corpora.

**Acknowledgement** This study is supported by the National Institute of Health (NIH) NIGMS (R00GM135488).

**References**

1. Siegel RL, Miller KD, Wagle NS, Jemal A. Cancer statistics, 2023. Ca Cancer J Clin. 2023 Jan 1;73(1):17-48.
2. Gutierrez BJ, McNeal N, Washington C, Chen Y, Li L, Sun H, Su Y. Thinking about gpt-3 in-context learning for biomedical ie? think again. arXiv preprint arXiv:2203.08410. 2022 Mar 16.
3. Moradi M, Blagec K, Haberl F, Samwald M. Gpt-3 models are poor few-shot learners in the biomedical domain. arXiv preprint arXiv:2109.02555. 2021 Sep 6.
4. Yang W, Soares J, Greninger P, Edelman EJ, Lightfoot H, Forbes S, Bindal N, Beare D, Smith JA, Thompson IR, Ramaswamy S. Genomics of Drug Sensitivity in Cancer (GDSC): a resource for therapeutic biomarker discovery in cancer cells. Nucleic acids research. 2012 Nov 22;41(D1):D955-61.